\documentclass[a4paper,conference]{IEEEtran}
\ifCLASSINFOpdf
  \usepackage[pdftex]{graphicx}
\else
\fi
\usepackage{amsmath}
\begin{document}
%
\title{Generalization of Fitness Exercise Recognition from Doppler Measurements by Domain-adaption and Few-Shot Learning}


\author{Biying Fu$^{1,2}$, Naser Damer$^{1,2}$, 
Florian Kirchbuchner$^{1,2}$, Arjan Kuijper$^{1,2}$\\
$^{1}$Fraunhofer Institute for Computer Graphics Research IGD,
Darmstadt, Germany\\
$^{2}$Mathematical and Applied Visual Computing, TU Darmstadt,
Darmstadt, Germany\\
Email: {biying.fu@igd.fraunhofer.de}
}


\maketitle

\begin{abstract}
In previous works, a mobile application was developed using an unmodified commercial off-the-shelf smartphone to recognize whole-body exercises. The working principle was based on the ultrasound Doppler sensing with the device built-in hardware. Applying such a lab-environment trained model on realistic application variations causes a significant drop in performance, and thus decimate its applicability. The reason of the reduced performance can be manifold. It could be induced by the user, environment, and device variations in realistic scenarios. Such scenarios are often more complex and diverse, which can be challenging to anticipate in the initial training data. To study and overcome this issue, this paper presents a database with controlled and uncontrolled subsets of fitness exercises. We propose two concepts to utilize small adaption data to successfully improve model generalization in an uncontrolled environment, increasing the recognition accuracy by two to six folds compared to the baseline for different users.
\end{abstract}


%
\IEEEpeerreviewmaketitle

\section{Introduction}

Human activity recognition (HAR) covers a wide range of application areas. Understanding human actions in daily living enables application designers to build assisting smart home applications for elderly care \cite{chen2011knowledge}, security applications with video surveillance \cite{10.1007/978-3-540-88458-3_88}, applications for Quantified-self \cite{Sundholm.2014,fu2020exertrack}, or associate physiological signals with emotions \cite{kim2004emotion} to build interactive applications. HAR is the key to enable human-centered application designs and natural interaction in a smart environment \cite{DBLP:journals/access/FuDKK20}.

However, human motion is highly complex and possesses a high degree of freedom. This is expressed with the term user-diversity. Learning a generalized model for all possible variations of a human motion is very challenging. Training a model on limited amount of individuals under constrained environment often leads to a large performance drop when applying the model on individuals/environments disjoint from the training data. This reduction in performance originates from the large variations between the controlled training data and the real-world application scenarios. It is caused by the diversity and complexity of the users actions, the device hardware or other environmental variations. However, all possible reasons lead to a degradation in the usability of the proposed application. One possible solution is to reduce the inherent difference between the development dataset and the real-world dataset by making the development data resemble the real-world data. However, due to the diversity in the real-world applications, there is no generalized model that is applicable in all possible situations.    

This work addresses sport exercise recognition from a stationary smartphone using the Doppler measurement. We propose a set of methods to improve the generalization of a pre-trained model (trained on controlled data) to scenarios containing a combination of unseen environments, individuals, and devices. To achieve this, we propose and investigate two concepts, along with a clear baseline that demonstrate the generalization problem.  We have developed a mobile application that aims at collecting data that is used to deal with this challenge. Our application is based on the built-in hardware of a commercial smartphone to measure whole-body exercise activities. The main contribution of this work is grouped as follows: 

\begin{itemize}
    \item A novel database for investigating micro-Doppler motion in relation to whole-body exercise data with built-in smartphone hardware. The database contains sessions in controlled environment, as well as a disjoint subset containing variations of environments, individuals, and devices.
    \item Propose and adapt two concepts (with variations) to improve the recognition generalization. These concepts are based on domain adaptions, as well as few-shot learning. Both concepts proved to enhance the generalizability on data variations in comparison to a clear baseline.
\end{itemize}

The structure of the paper is organized as follows: in section \ref{ch:related} we provide current researches on the topic of finetuning approaches with the focus on model generalization to fit new data and categorized these approaches under two main categories (retrain required and not). In Section \ref{ch:database} we introduce one of the main contributions of this paper by presenting our collected database. We first motivate the need for such a database. We then introduce the sensing principle and describe the details about the database and what it enables us to study. In section \ref{ch:methods}, we propose the baseline model and the new approaches targeting our problem, under two main concepts. Section \ref{ch:evaluation} introduces evaluation results of our proposed individual approaches, along the baseline. We further discuss the advantages and disadvantages of certain methods and provide some guidelines in design choice for such an application. Finally, we conclude our work in Section \ref{ch:conclusion} and provide relevant future research directions.

\section{Related Works}
\label{ch:related}
Andrew Ng once stated \cite{ng2017machine}, that the most frequent fail of inference model in reality is the inherent difference between your development set and test set. To overcome this problem, effort can be put into developing highly-representative development/training databases. Though, it is impossible to make the development set identical to the test set (application scenario), due to the large variety and complexity in human activities. Common methods to adapt the pre-trained model on individual new data samples without enforcing much restrictions on the development set is thus desirable. We distinguish between two main categories: with and without retraining the base model to adapt to new data.

\paragraph{Retraining of the Base Model}
Domain adaptation builds on transferring knowledge from similar domains to cope with unknown target domains. This method is especially useful, when you do not have enough annotated datasets for the particular problem at hand. The goal is to extract knowledge from related, known datasets, and use this knowledge to learn the new task at hand.

Wang \cite{wang2018deep} benefited from using similar labeled source domain data to annotate the target domain that has only a few or even no labels. They evaluated their approach on acceleration dataset from different body positions as different domains. To alleviate the problem of negative knowledge transfer, they proposed an unsupervised similarity measure to choose the \textit{right} source domain with respect to the target domain. Khan and Roy et al. \cite{khan2018scaling} proposed a CNN based transductive transfer learning model to adapt action recognition classifiers trained in one context to be applicable to a different contextual domain. The limitation is that the set of activities being monitored is the same in both context domains, as they are transferring knowledge from individual convolutional layers. Evaluated on their acquired smartphone and smartwatch acceleration dataset on 15 users and 8 activities, they demonstrated the ability of their proposed methods on transferring knowledge from smartphone to smartwatch domain and vice versa. By incorporating a small amount of target labels, they were able to further increase the performance. 

These methods pose less constraints on the target domain, however are difficult to train, as the knowledge transfer is solely based on the source domains. Thus the choice of the appropriate source domain is critical for the performance of the inference model on the unknown target domain. A retraining is required to relate the source domain to the new target domain due to knowledge transfer.

\paragraph{No Retraining of the Base Model}

Few-shot learning is currently an active research field in machine learning. The ability of deep neural networks to learn complex correlations and patterns from a vast dataset is proven. However, current deep learning approaches suffer from the problem of poor sample efficiency. To make a model learn on a new class, sufficient amount of labeled samples from this class is required to avoid overfitting.

Few-shot learning methods increase the model generalizability with limited data. They have been mostly used in image classification tasks \cite{snell2017prototypical,koch2015siamese}. Feng \cite{feng2019few} recently applied few-shot learning-based classification on human activity recognition tasks. They applied a deep learning framework to automatically extract features and perform classification. However, instead of transferring knowledge in a common feature space, their proposed method tends to perform knowledge transfer in terms of model parameter transfer. Based on two benchmark datasets, they evaluated their proposed technique. A metric to measure the cross-domain class-wise relevance is introduced to mitigate the challenging issue of negative transfer. These datasets consist of only sparse sensory input, mostly acceleration sensor data attached to different body parts. 

We consider this category to be the most realistic in designing applications for human activity recognition tasks with sensory data. Since we can not adopt to the complexity and diversity of all persons actions during the training phase, we need the network to have the ability to adapt to individual users by introducing only a small amount of this users data to optimize the trained network. Few-shot classification methods are methods that can be leveraged to unseen classes, even when less labels from these classes are available without retraining the base model.

Exercise detection on personal devices is often applied to track daily ambulation activities \cite{kwapisz2011activity,ravi2005activity}. For tracking of more stationary activities involving whole-body interaction, a remote system is better than wearing a smartphone on the body, as the detection is more unobtrusive. However, the complexity and diversity in human action makes it difficult to develop one single system to fits all situations. To overcome this issue, it requires more advanced machine learning methods to improve the model generalization. 

\section{Ultrasonic smartphone exercises: sensing and database}
\label{ch:database}
\begin{figure*}
\centering
\begin{tabular}{cccc}
   \includegraphics[width=25mm]{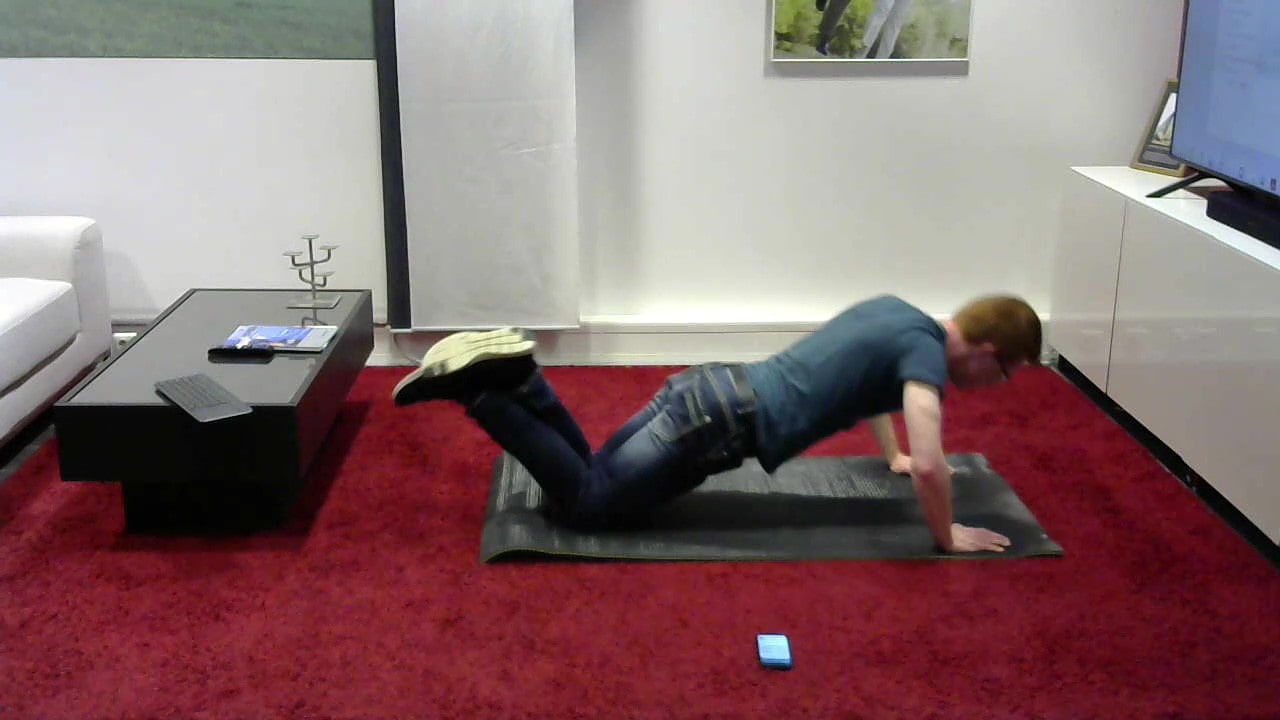} &   \includegraphics[width=25mm]{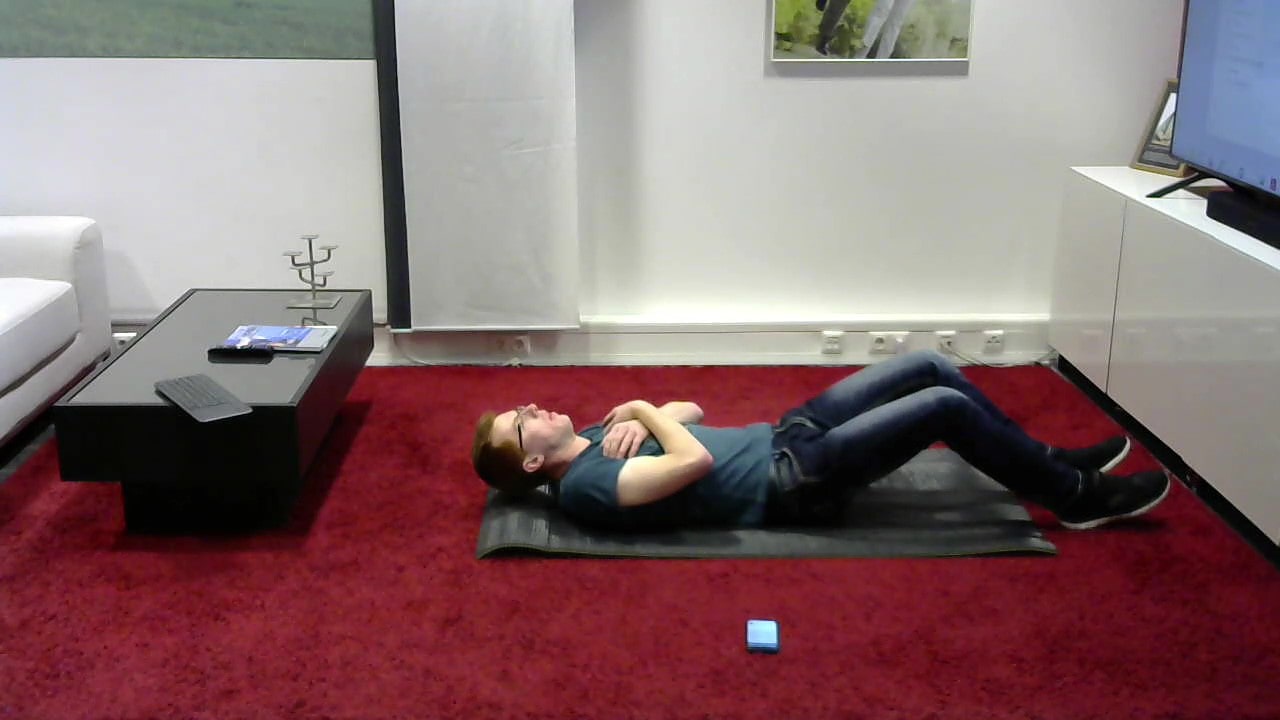} &   \includegraphics[width=25mm]{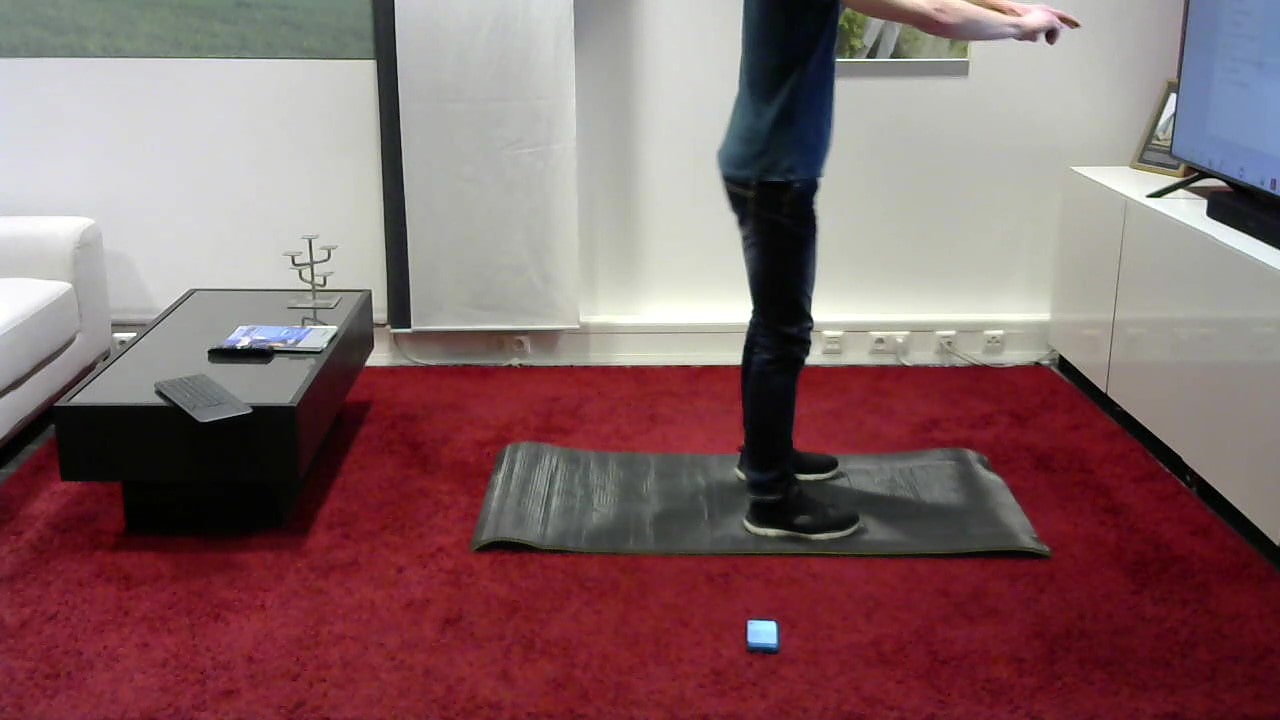} &   \includegraphics[width=25mm]{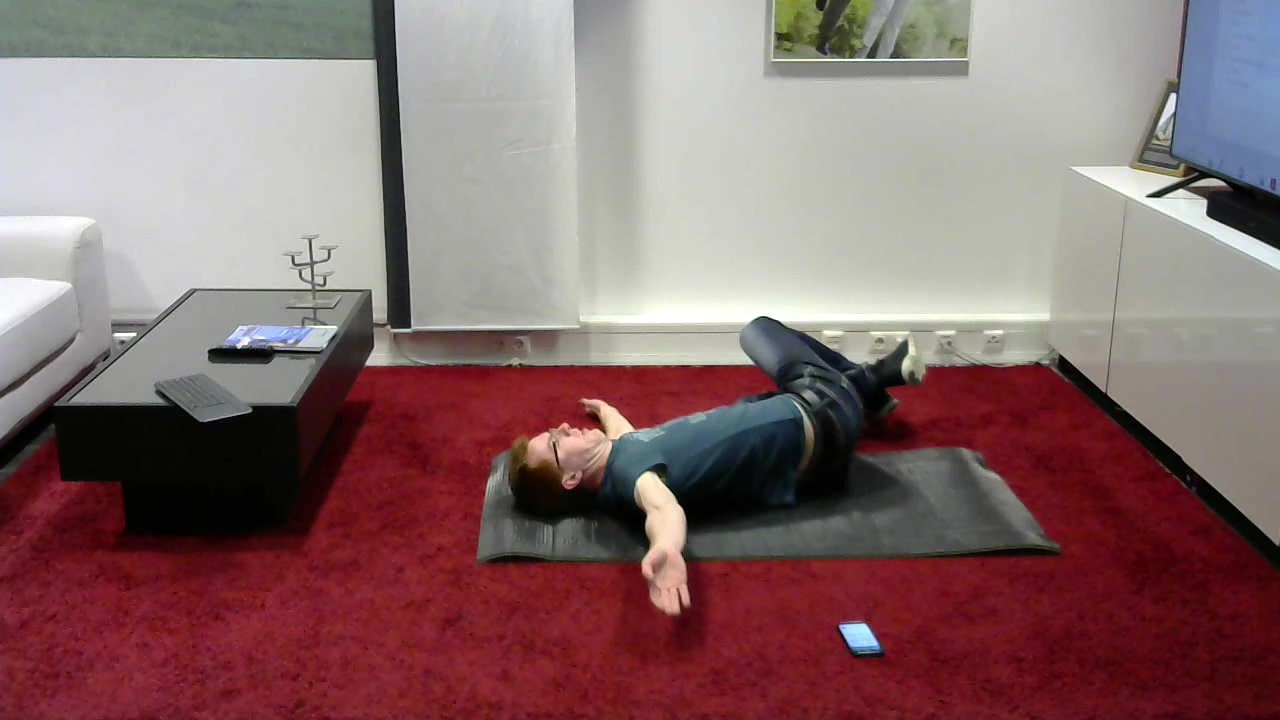} \\
    push-up & sit-up & squat & segmental rot. \\[6pt]
    \includegraphics[width=25mm]{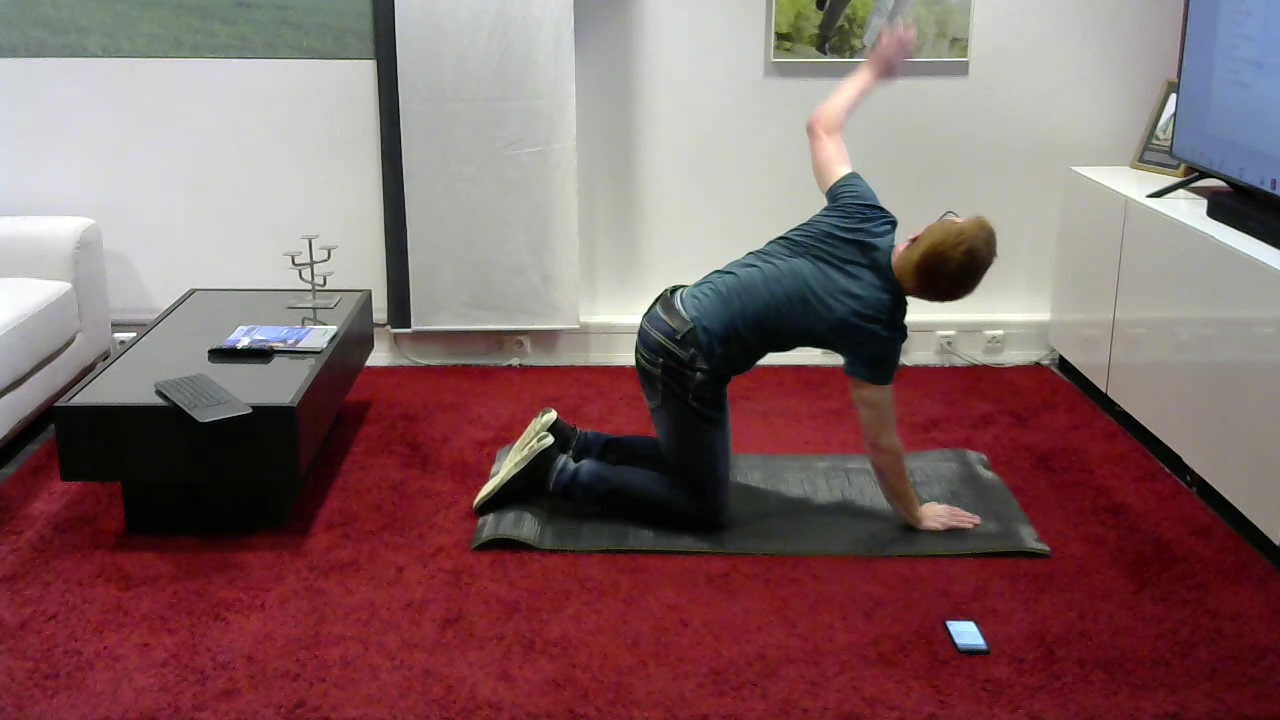} &   \includegraphics[width=25mm]{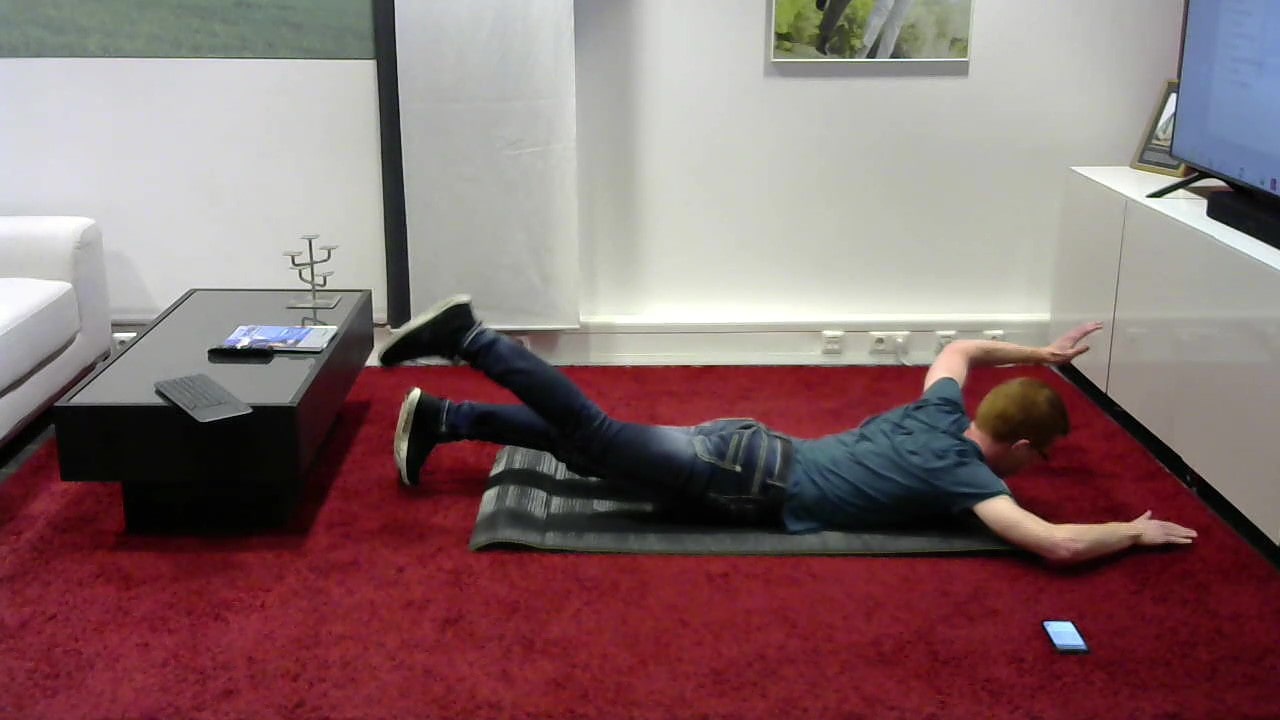} &   \includegraphics[width=25mm]{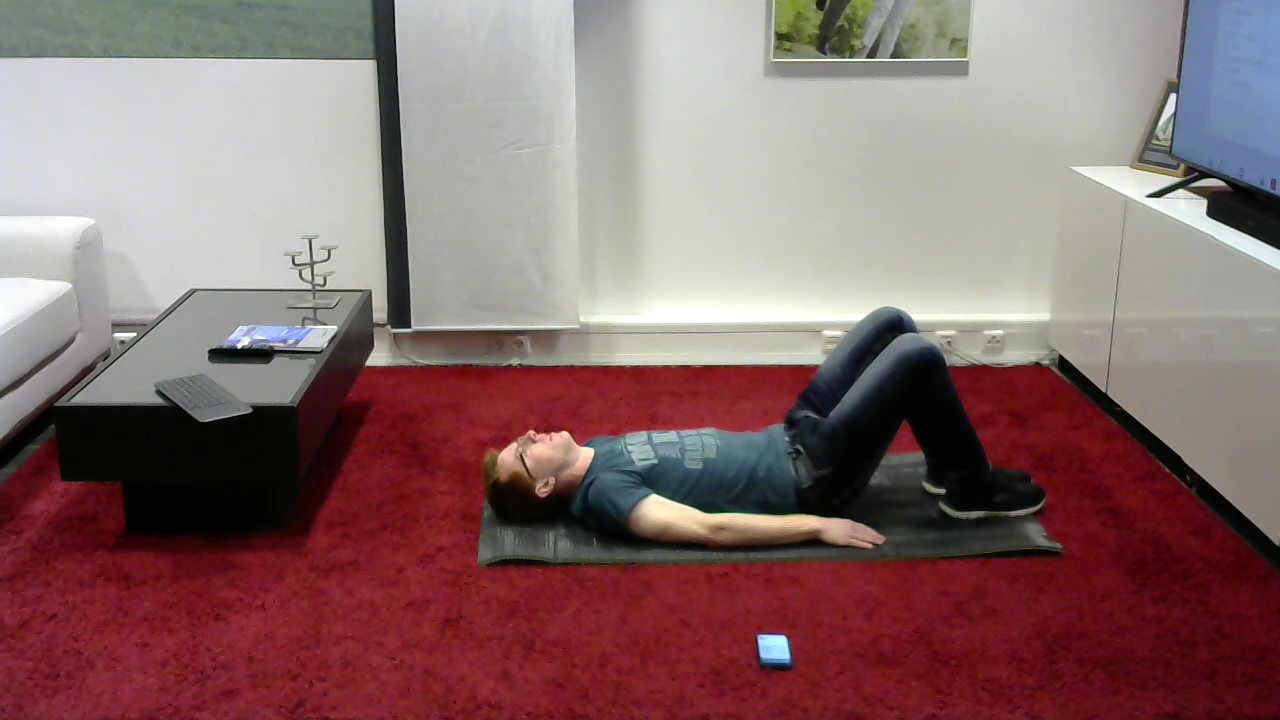} &   \includegraphics[width=25mm]{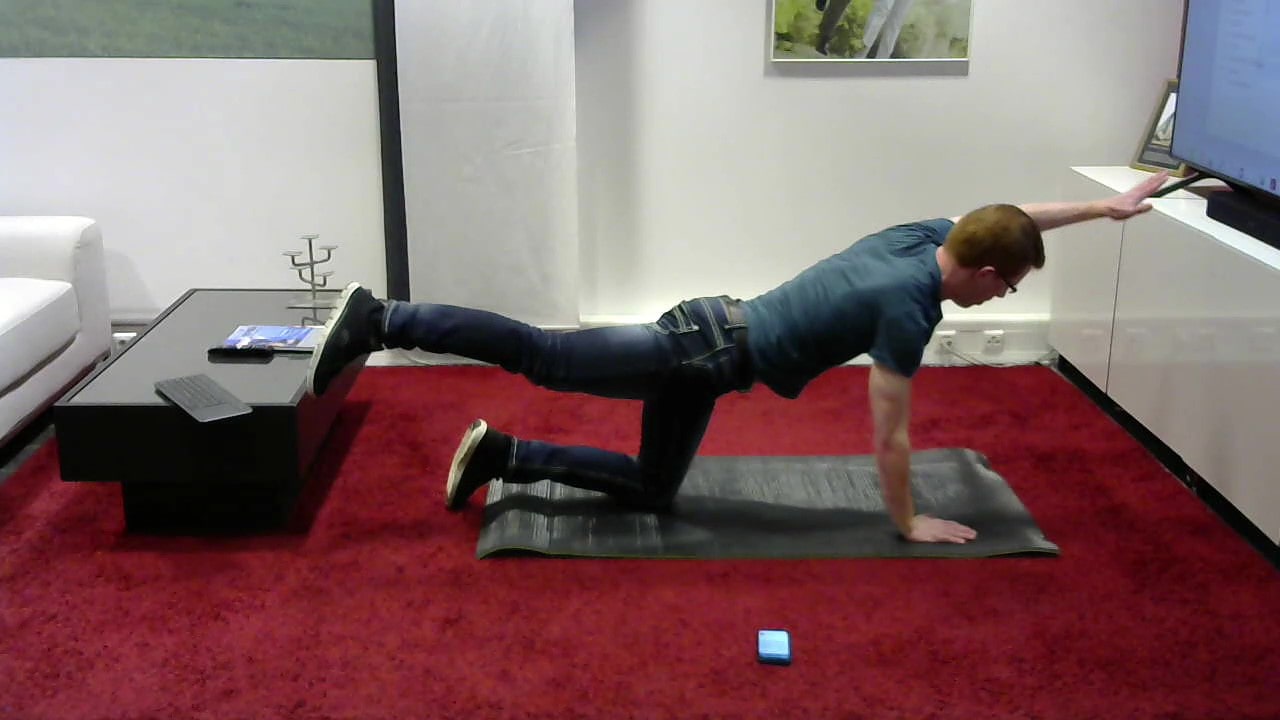} \\
    trunk rot. & swim & bridge & quadruped \\[6pt]
\end{tabular}
\vspace{-3mm}
\caption{Figure shows the eight workout exercises we present in our novel database. The smartphone is placed roughly 50\,cm away from the body of the performing user.}
\label{fig:exercise_classes}
\end{figure*}

In this paper, we contribute a database collected with built-in hardware of commercial smartphones. This database deals with exercise data using Doppler sensing by utilising the smartphone as a sonar device. Using smartphone to collect activity data is not novel per se. Most existing databases are, however, focused on acceleration data using the smartphone as a wearable device. Popular databases are for example provided in the works \cite{anguita2013public,van2011human,kwapisz2011activity}. Though tracking and recognizing for various aerobic training exercises are popular, there exists limited research on recognizing more stationary exercises, such as strength-based training without the use of wearable. These exercises are essential for rehabilitation purposes \cite{U.S.DepartmentofHealthandHumanServices.2008}. Typically they are even harder to track, as they rely on coordinated movement of specific body parts.  Instead of building customized hardware designs such as in \cite{Sundholm.2014, fu2020exertrack} to target these exercises, we leveraged the existent infrastructure of a commercial mobile device to build an ubiquitous application. The set of stationary and strength-based exercises are depicted in Figure \ref{fig:exercise_classes}. 

\subsection{Sensing Principle}
By emitting a 20\,kHz continuous audio signal from the built-in speaker, we turned the commercial smartphone into an active sonar device. The echo encoded with the Doppler modulation is received from the device internal microphone. Doppler measurement allows us to catch relative movement in close range to the sensing device. A positive Doppler is received with a relative speed towards the device and a negative Doppler vice versa. The device speaker can typically emit tones in the range of 18-22\,kHz on a commodity audio system without performance degradation. Thus, we can detect a one-sided Doppler speed up to $17.4\frac{m}{s}$ (2.05\,kHz). To extract the relative movement from Doppler signal, a Short Time Fourier Transformation (STFT) \cite{durak2003short} is used to convert the continuous time signal to the spectrum domain of interest. The parameter of STFT determines the resolution in time and frequency domain. The selected frequency resolution corresponds to a relative speed of $3\frac{cm}{s}$ (3.6\,Hz) with a time resolution of 46.5\,ms. A sliding window approach of 6\,s windows and an overlap of 50\,\% are chosen to prepare data samples used to train the classification task. To further reduce the computational effort, we restrict the spectrum amplitudes within frequency band between 19.5 kHz to 20.5 kHz, as other signals beyond the motion information are irrelevant. A typical \textit{push-up} exercise takes around 2-3\,s each \cite{yoo2014effect}, depending on the individual fitness. We conclude that with the current system setting, we are able to resolve both the slowest and fastest movement of the targeted exercise set. 

\vspace{-1mm}
\subsection{Database}
\vspace{-1mm}

This database allows us to study the body motion in relation to Doppler profiles from built-in hardware of various commercial smartphones. The effect of fine-grained movements from both limbs and arms cause micro-Doppler patterns in addition to the main Doppler reflection. Studying these micro-Doppler events enhances the ability of recognizing more complex and naturalistic human activities including whole-body interactions. To the best of our knowledge, there does not exist such a database so far. Due to the similarity of ultrasound sensing and electromagnetic waves in physical characteristics, this database can also be leveraged for machine learning practitioners to design radar-based applications and gain useful insight without the additional cost of customized hardware.

The presented database consist of two different setups, in order to investigate the effects of various methods on improving model generalization for different data distributions. Data of the first setup is called Lab-Data. The Lab-Data consists of data collected in a laboratory setup as depicted in Figure \ref{fig:exercise_classes} from 14 individuals. The group consists of 4 females (157\,cm-172\,cm) and 10 males (172\,cm-193\,cm). The affinity towards sport exercises varies across the test participants. The built-in microphone of the sensing device is placed 50\,cm apart, facing the exercising individuals on the floor, aligned with the hip as depicted in Figure \ref{fig:exercise_classes}. For each individual, two separate sessions were collected with 10 repetitions of each exercise class. Left and right variations for exercises such as \textit{segmented rotation, trunk rotation}, and \textit{quadruped} are counted as one repetition. \textit{Swim} is performed in average for 30\,s in each session to reach similar time duration comparable to the other exercise types. In order to collect the micro-Doppler motions from the arms, the device is aligned with the shoulder for \textit{swim} and \textit{trunk rotation}. The duration of each session is approximately 7-9 minutes in average. The smartphone used for data acquisition has the brand Samsung Galaxy A6 (2018) and the placement of the sensing device to the exercising body is constrained to the same position for all participants. 

\begin{figure*}
\centering
\begin{tabular}{cccc}
   \includegraphics[height=2.3cm, angle =90]{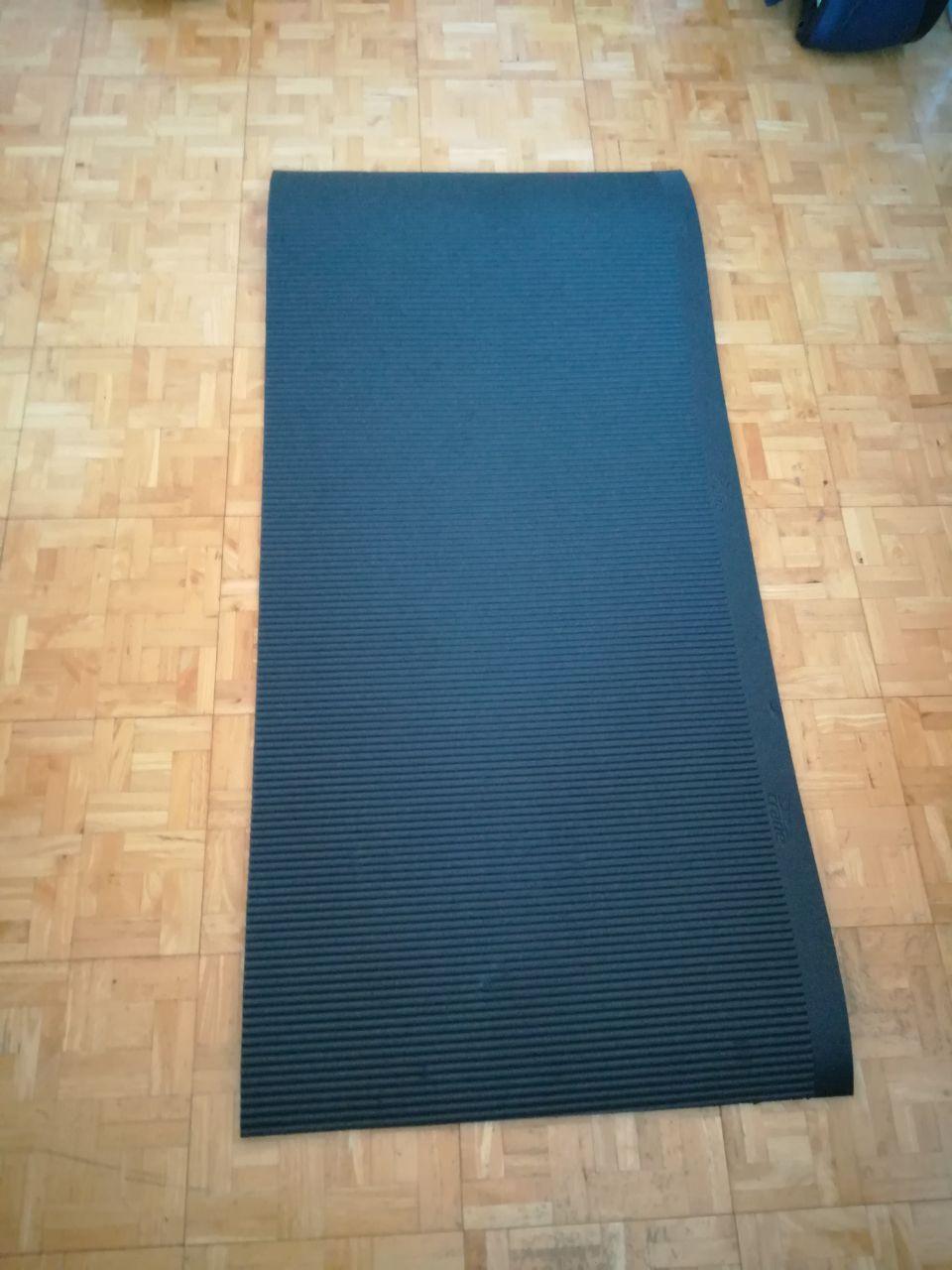} &   
   \includegraphics[height=2.3cm, angle =90]{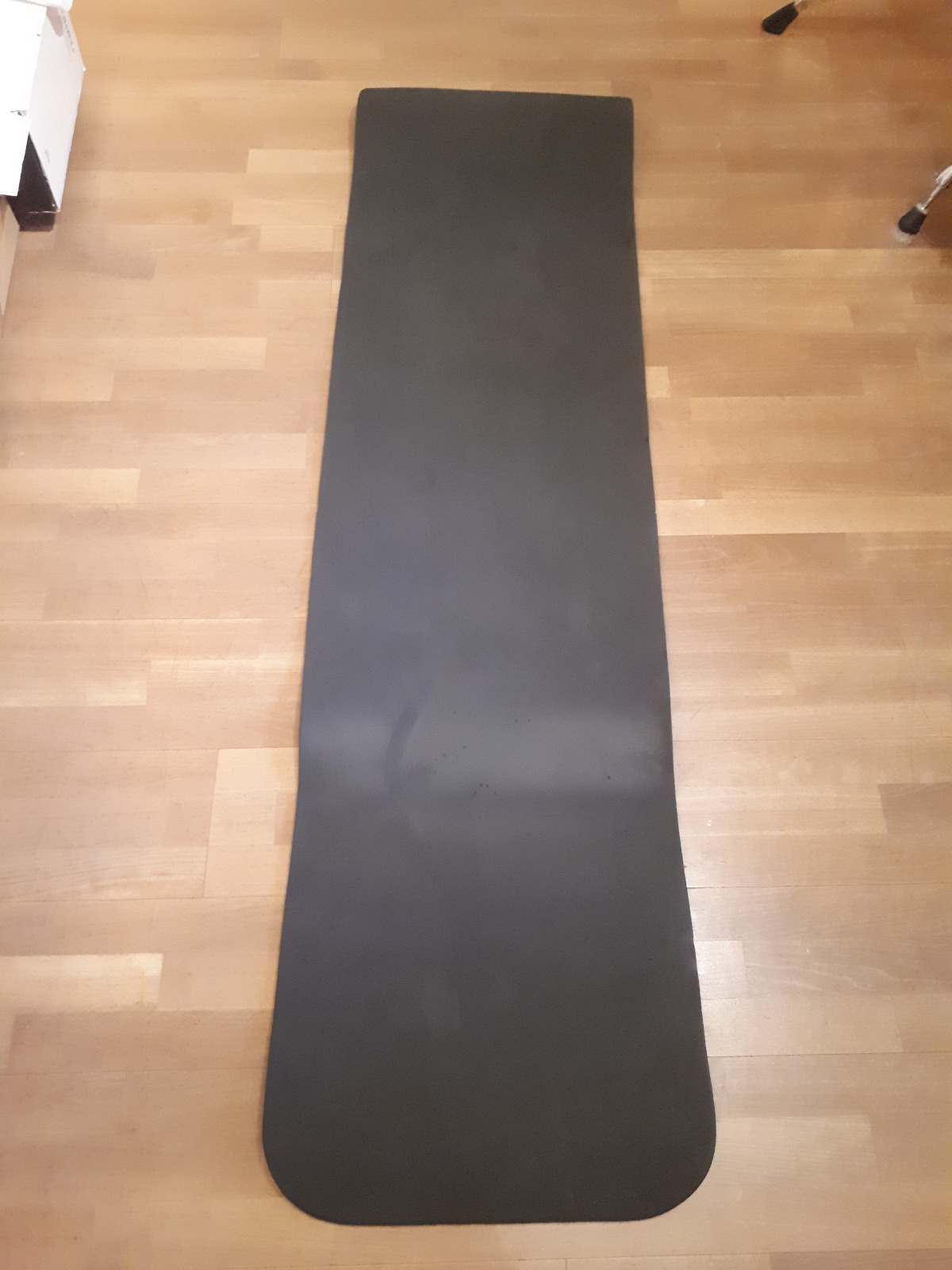} & 
   \includegraphics[height=2.5cm, angle =90]{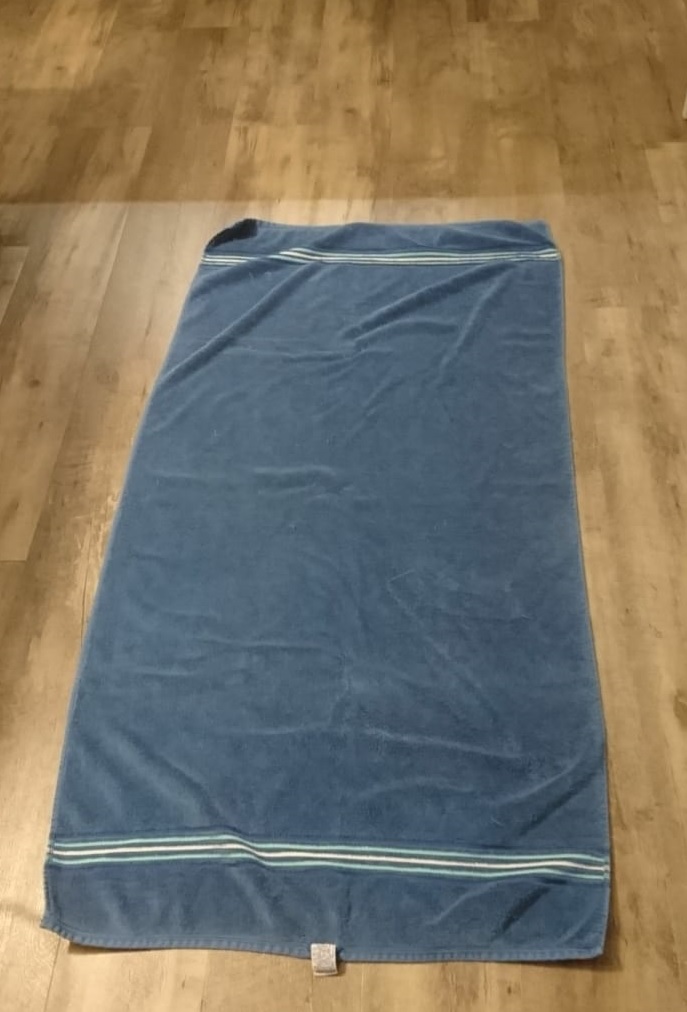} &   
   \includegraphics[height=2.4cm, angle =90]{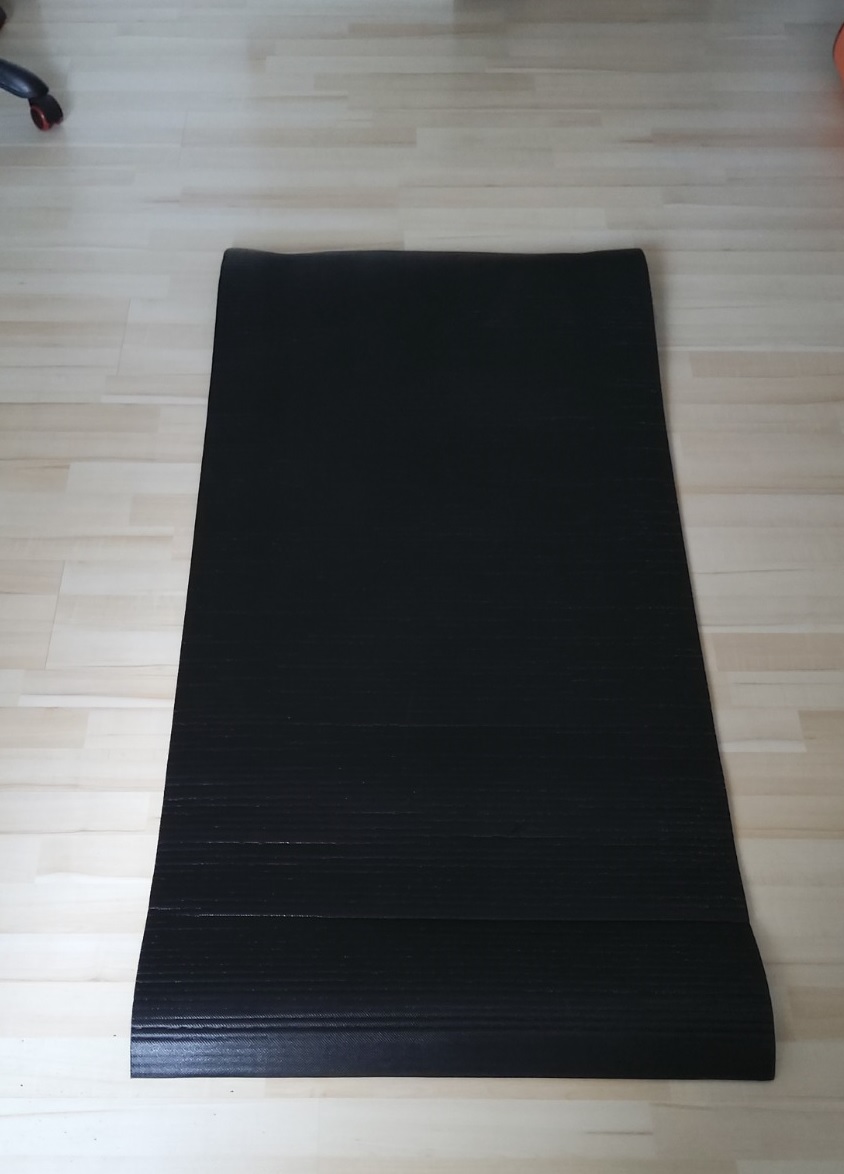} \\
   Environment 1 & Environment 2 & Environment 3 & Environment 4\\[6pt]
\end{tabular}
\vspace{-3mm}
\caption{Illustration depicts the four different data acquisition setups from the individual test participants.}
\label{fig:apartment}
\vspace{-3mm}
\end{figure*}

The goal of the second setup is to be leveraged on testing various finetuning approaches, as these data are collected under individual, different hardware, and uncontrolled environments independent of the Lab setup. This part of data is called the Uncontrolled-Data. It consists of data collected from five different individuals. Due to logistic and privacy constrains related to the experimental setup, the second setup contains a smaller number of participants compared to the Lab-Data. The hardware device is not limited to the smartphone used in the Lab-Data. Each individual was asked to collect eight individual sessions distributed over several days in their familiar surroundings and without any supervision. Each session has a comparable length to the collected data from the Lab-Data. Some general statistics about the participants and the hardware devices used, are listed in Table \ref{tab:population}.
\begin{table*}[]
    \centering
    \begin{tabular}{c|c|c|c|c|c}
    \hline
         Participant & Sex & Height &  Exercise Frequency & Device & Location  \\
         \hline
         P1 & male & 180\,cm & Frequent & SONY Xperia XZ2 Compact & Environment 1 \\
         P2 & male & 181\,cm & Frequent & Samsung Galaxy A5 (2017) & Environment 2 \\
         P3 & male & 181\,cm & Frequent &  SONY Xperia Z5 Dual & Environment 3 \\
         P4 & male & 182\,cm & Frequent & Samsung Galaxy A6 (2018) & Environment 4\\
         P5 & female & 168\,cm & Less Frequent &  SONY Xperia XZ2 Compact & Environment 1\\
         \hline
    \end{tabular}
    \caption{Description of Software and Hardware Setups for the Uncontrolled-Data.}
    \label{tab:population}
\end{table*}
The data acquisition app is installed on the individual mobile device. The participants from the uncontrolled setup were asked to collected data from their home environment to simulate the real-world scenarios. Figure \ref{fig:apartment} illustrates the different data acquisition environments that affect the signal strength of the underlying hardware device. In contrast to the Lab setup, the other apartments all have wooden floors which makes the back reflected signal strength stronger compared to the Lab setup.   

This paper aims at investigating methods to improve the generalizability of pre-trained models on new individuals under more realistic conditions. Based on the underlying method, we split the data as follows: 

\begin{itemize}
    \item Basic training data contain all individuals and sessions of the Lab-Data.
    \item Subject development data contains 4 sessions (out of 8) of each of the users in the uncontrolled setup, the Uncontrolled-Data.
    \item Testing data contains the 4 sessions (out of 8) of each of the users in the uncontrolled setup, the Uncontrolled-Data, that were not used in the Subject development data.
\end{itemize}

\section{Methods}
\label{ch:methods}
This paper is motivated by our observations from our previous experiences in realistic use of mobile device to detect workout exercises without using additional external hardware \cite{fu2020_ultrasense}. However, previous work faced a major usability challenge as it did not adapt well on individuals and environments unseen in the training phase. This issue is the main target of the methods presented in this paper. To state the problem, we observed the input signal of the same participant performing the same set of physical activities under two different environments with the same hardware and sensor position. Despite the overall speed and appearance remain similar, the strength and noise embedded reveal a strong difference in both settings. We noticed a strong decay in signal power due to the material-dependent attenuation of the transmit power.

Realistically, we can not train a classifier adapting to every possible sensing environment, unless our training data unrealistically contain unlimited variations. The quality of hardware devices integrated in the smartphone may also introduce strong variations in the signal power. But the basic physical characteristics remain. To adapt to new, real-world circumstances, we need to individually finetune the trained model. In this paper, we investigate several approaches to improve the model generalizability given limited individual data.
\vspace{-2mm}
\subsection{The baseline method}
\vspace{-1mm}
The base inference model is built using a stacked bidirectional LSTM network. To baseline our proposed solution, we need to demonstrate the exercises detection performance when the uncontrolled environment is not considered. The choice of using this sequence model is due to its ability to consider the global and sequence structure within a sample time window. The architecture of this sequence modeling network consists of 2 stacked bidirectional LSTM layers with 128 hidden nodes in each LSTM cell. For each input node, a slice of spectrogram with the frequency bands (ranging from 19.5\,kHz to 20.5\,kHz) from a time step resolution (46.5\,ms) is provided to the network. The bidirectional structure permits the network to look forward and backward in time to extract fine-grained sequence information from the spectrum domain. 

An 2-D instance normalization layer is applied on the sample spectrogram prior to the network input in order to reduce the hardware specific power dependencies. The number of output classes are nine that includes the eight true activity classes with the additional \textit{none} class describing all the transitions and noisy samples between two successive action classes. The learning rate is set to 0.001 and the Adam Optimizer is used to optimize the network parameters. Cross entropy loss is used as the cost function to minimize the loss of the misclassification error from the training samples. Batch-wise training is used, while in each batch, 15 samples of each class are randomly selected from the training data to construct similar training procedures compared to other network structures. 

The training contains the Lab-Data only. Subject development data with 4 sessions from the Uncontrolled-Data is used in the validation stage, while the 4 sessions of the remaining Uncontrolled-Data is used in the test phase. Batch-wise approach is applied, while each batch contains 15 samples randomly selected from each class. 

\vspace{-2mm}
\subsection{Our proposed method 1: Domain adaptation}
\vspace{-1mm}

To improve the model generalization ability, the first method we propose is from the domain adaptation (DA) network family. DA is commonly used to transfer knowledge from labeled source domain to target domain where data is unlabeled or only partially labeled. In our case, the source domain refers to the Lab-Data, while the target domain refers to the Uncontrolled-Data collected by the different individuals under changed conditions. The aim of such an adaption network is to make the distribution of the source and target embeddings in the common embedding space more similar, such that the classification network works well on both domains. By applying this architecture, we aim at adopting the base feature extractors to be sensitive to deterministic features from the Uncontrolled-Data domain with the knowledge draw from the source domain. In such a way, the classification performance does improve for the uncontrolled setup.

\begin{figure}
    \centering
    \includegraphics[width=0.80\linewidth, trim={0 -2.5cm 0 0}]{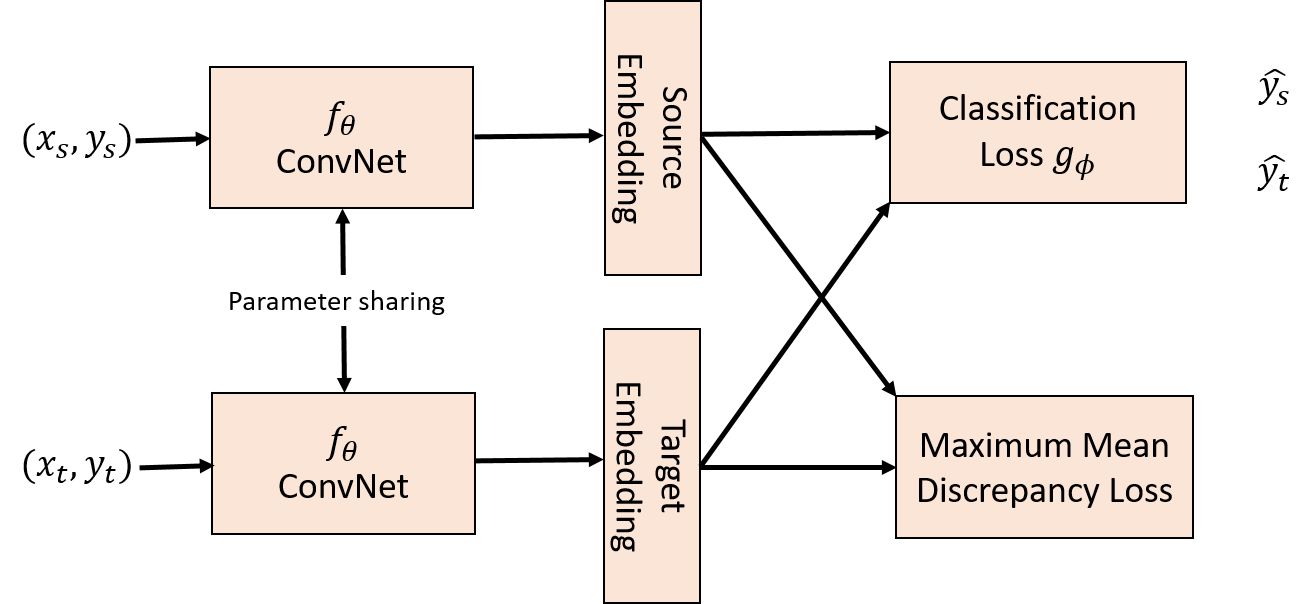}
    \vspace{-0.7cm}
    \caption{depicts the domain adaptation model. Source and target data are projected into the same embedding space using the common base ConvNet. Adaptation is realized by minimizing the classification losses for source and target and reducing differences in both feature distributions.}
    \label{fig:model_retrain}
    \vspace{-3mm}
\end{figure}

The metric to measure the similarity of both distributions is based on the maximum mean discrepancy (MMD) on the final feature level. The model architecture of the adaptation network is depicted in Figure \ref{fig:model_retrain}. Common approaches using domain adaption do not require to include target labels. However, without any label information from the target domain, the knowledge transfer does not work well on the targeted use-case, as both domains are quite dissimilar. In order to improve the knowledge transfer characteristic, we included partial labels (50\,\%) from the subject development set of the Uncontrolled-Data to increase the performance on feature level adaptation. An instance normalization layer is used prior to the ConvNet to mitigate the hardware dependent effects of the transmit power from different smartphone models. 

A base ConvNet is used to extract common features from source and target domain. The ConvNet structure consists of 4 successive convolutional layers, each followed by a batch normalization layer, leaky rectified linear unit (ReLU) activation layer with the negative slope coefficient of 0.2 and a max pooling layer to reduce the input dimensions. The successive layers increased the number of filters from 32-32-64-64. The same ConvNet structure is used for all networks as the feature extraction component in this work. The embeddings in the embedding space are used to minimize the classification loss of the source and target domain. The objective is to minimize the combination of three different losses as given in Equation (\ref{eq:loss_cda}) : both cross entropy losses from the source and target domain classification, and the MMD loss of the feature embeddings originated from the source and target domain in the same embedding space.
\begin{equation}
     \mathcal{L} =  \mathcal{L}(g_{\phi}(f_{\theta}(\textbf{x}_s)),y_s) + \mathcal{L}((g_{\phi}(f_{\theta}(\textbf{x}_t)),y_t) + \mathcal{L}_{MMD}
\label{eq:loss_cda}
\end{equation}

The network is trained with 100 epochs. Each batch composites of 15 samples from each of the 9 classes. Adam optimizer with a learning rate of 0.001 is set to learn the network hyperparameters. Since the base feature extraction network needs to be optimized on both data domains, a retraining of the base model is required. We applied the Lab-Data as the source domain and the uncontrolled subject development set as the target domain. The adapted model is evaluated on the testing data of the uncontrolled setup.

This method is good to adopt the feature extraction layers to work for features from different domains. A negative aspect is that if both domains differ too much, it could lead to a negative adaptation and causing the performance on the source data to decrease. To address this issue, we present the following methods which do not need to retrain the pre-trained inference model to finetune to new individuals. We benefit from a few labels of the new datasets to label this unknown dataset.

\vspace{-1.5mm}
\subsection{Our proposed method 2: Few-shot classification}
\vspace{-1mm}

This section deals with three methods from the research domain of the few-shot classification learning. The network is trying to learn common features within a subset of tasks without retraining. 

\paragraph{F1: Siamese Network with Few-Shot Classification}
\label{sub:siamese}

\begin{figure}
    \centering
    \includegraphics[width=0.8\linewidth]{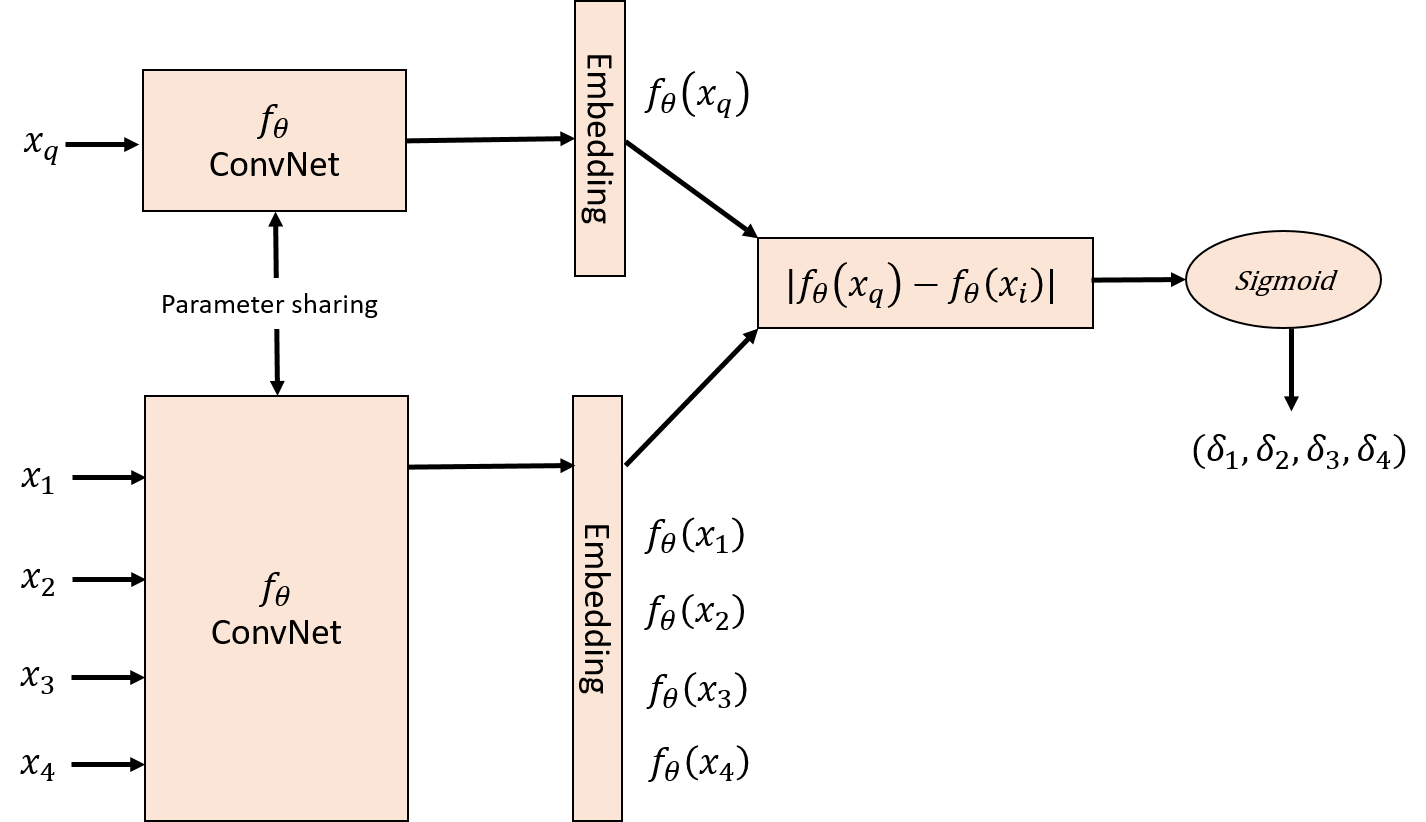}
    \caption{The Siamese network for few-shot classification task is to learn the optimum separability for the multiclass classification problem. The ConvNet structure is used to generate feature embeddings. A distance measure is calculated for the query sample with all possible support classes. The output label is the highest similarity score within the multiple classes.}
    \label{fig:siamese}
    \vspace{-3mm}
\end{figure}

The Siamese network consists of two identical feature extraction base networks with shared weight parameters. The learned feature embeddings from both inputs are then compared with each other to form a similarity score. Commonly it is used for verification tasks and the score indicates how similar two input samples are. Instead of the verification task as performed in common Siamese networks, we extended it for the few-shot classification task. In contrast to the domain adaptation method, this method can train on disjoint data. The Uncontrolled-Data is not used in the training stage. The working principle is depicted in Figure \ref{fig:siamese}. Our network structure aims at learning the optimum separation between all multiple classes at once. 

During training time, each training batch consists of 15 query samples from each class. The support set consists of 5 support samples from each class. They are fed through the feature extraction network ConvNet $f_{\theta}$ to get the embeddings. In case the support of each class is larger than one sample, a mean embedding is calculated to reduce the computational complexity. A similarity score is determined for the query sample embedding $f_{\theta}(\textbf{x}_q)$ and the individual support class embeddings $f_{\theta}(\textbf{x}_i)$, where $i=1..C$ and $C$ represents the number of classes. To determine the similarity score, an euclidean distance vector between the feature embeddings is calculated by $\left|f_{\theta}(\textbf{x}_q) - f_{\theta}(\textbf{x}_i)\right|$. This distance measure is further fed through a dense layer to learn the final similarity score $\delta_i=\sigma(\phi(\left|f_{\theta}(\textbf{x}_q) - f_{\theta}(\textbf{x}_i)\right|))$. This parametric dense layer is optimized by a pair of input samples every time step. Based on the similarity measure between the query sample and all other support classes $(\delta_1,\delta_2,\delta_3,...,\delta_C)$, the objective of the Siamese network is to maximize the maximum likelihood estimation of data pairs or equally to minimize the negative log likelihood cost function.

Adam optimizer is used to train the network parameter with a learning rate of 0.0005. 500 Epochs were performed during the training phase with 9-ways-5-shots. Fifteen query samples each class are used to build the query set in each epoch. Nine classes are learned in each individual training task with 5 support samples from each class. In the evaluation phase, the number of supports each class can be tuned between 5 and 10 samples each. We perform 100 iterations to obtain the average accuracy on the test set. 

The similarity measure of the Siamese network here is learnt by a parametric dense layer, in the next two paragraphs we introduce two other alternatives from the few-shot classification task where a non-parametric distance metric-based learning is used in the classification stage. These methods are more robust against small difference in source and target domains, and thus more generalized.

\paragraph{F2: Prototypical Few-Shot Classification Network}
Inspired by \cite{snell2017prototypical}, we adopted the prototypical network as the second method (F2) in the few-shot classification problem. In comparison to the first few-shot method (F1) with Siamese network, this network is non-parametric in the classification stage. This approach is a feature metric-based learning approach. Instead of transferring network parameters, this learning approach is based on learning the similarity distance between the feature embeddings from new target data to the prototypes of the samples from the same setup. This method is similar to a clustering-based method to find the k-nearest neighbours used for the classification.

The negative log-likelihood (NLL) function is applied to the negative euclidean distance of the embedding vectors to each class center embeddings (prototype). The objective is to reduce the NLL loss of the query output to the true classification label. Adam optimizer is used to optimize the weights of the ConvNet hyperparameters. A learning rate of 0.001 is used to learn the hyperparameters. 500 Epochs were performed to train 5-ways-10-shots classification tasks. 15 query images each class are used to build the query set. In the evaluation phase, the number of class recognition task has increased to the total 9 classes. The number of supports from each class is varied between 5 and 10 samples each. We perform 100 iterations to obtain the average accuracy on the test set.

The disadvantage of using euclidean measure to express the similarity is that this measure is not bounded. In contrast to Siamese network, there is non parametric learning after the feature extraction ConvNet structure. The classification is based on the k-nearest neighbour approach from the class embedding centers. The last method uses the cosine similarity measure which is a bounded metric.

\paragraph{F3: Local Descriptor Correlated Few-Shot Classification Network}

The third method of the few-shot classification (F3) is inspired by Li \cite{li2019revisiting}. The author introduced a new distance metric to label the query sample. Instead of using image-level feature based measure, they introduced a local descriptor based image-to-class measure. We think this architecture will work on the 2D time-frequency spectrum, as this network is sensitive to local features within a global context. In contrast to object images, the Doppler spectrum contains less fine-grained contextual information, but local features caused by repetitive movements and micro-movements from limbs and arms are clearly observable. In this paper, we examine the local descriptor correlation from the structural information extracted from the micro-Doppler range caused by different whole-body activities. 

The local features from the ConvNet output is correlated with all other local descriptors of the support embeddings each class using a cosine similarity. This provides a locally feature-based image-to-class mapping. The advantage of the cosine metric is because the cosine distance measures the pattern similarity without being largely effected by the magnitude. The objective is to reduce the cross entropy loss for this multiclass classification problem. Adam optimizer with a learning rate of 0.001 is chosen. The parameters for the few-shot learning and the batch-wise composition are the same as in the method F2 for the prototypical network. 

In few-shot classification learning task, the training data can be disjoint of the test data. We used the Lab-Data to build the training tasks and used the subject development data as support set and the testing data from the uncontrolled setup to evaluate the model.

In this section, we introduced three methods from few-shot learning: (F1) Siamese, (F2) ProtoNet, and (F3) LocalNet. These methods are theoretically suitable to enhance the generalization for our use-case, since we want to mitigate the retraining phase for similar tasks. 

\section{Evaluation and Discussion}
\label{ch:evaluation}
The database details are introduced in Section \ref{ch:database}. In this section, we present and discuss the results of the proposed approaches in comparison to the baseline. We aim to optimize the trained model with Lab-Data under controlled condition to be generalizable to individuals under the uncontrolled setups. 

The results of the baseline model and models which require adaptation with the Uncontrolled-Data during the training phase are displayed in Table \ref{tab:retrain}. 
\begin{table*}
    \centering
    \begin{tabular}{l|c|c|c|c|c|c}
    \hline
         Method & \shortstack[l]{Ratio of labels used\\ from Uncontrolled-Data}  & P1 & P2 & P3 & P4 & P5\\
         \hline
         Baseline & - & 16.25\,\% & 46.35\,\% & 11.17\,\% & 25.73\,\% & 15.5\,\%\\
         \hline
        DA & 0\,\% & 35,20\,\% & 37,80\,\% & 20.14\,\% & 57.67\,\% & 38.04\,\% \\
        \hline
        DA & 50\,\% & 77.61\,\% & 85.39\,\% & 58.61\,\% & 78.86\,\% & 75.87\,\% \\
        \hline
        DA & 100\,\% & \textbf{87.13\,\%} & \textbf{98.14\,\%} & \textbf{84.10\,\%} & \textbf{76.92\,\%} & \textbf{96.85\,\%} \\
    \hline
    \end{tabular}
    \caption{The accuracy results of baseline method and methods with model retraining is depicted. The term $P_i$ indicates the ID of the participants. Enabling the knowledge from the Uncontrolled-Data to modify the base feature extraction increases the performance on individual finetuning.}
    \label{tab:retrain}
    \vspace{-3mm}
\end{table*}

\vspace{-1mm}
\subsection{Baseline results}
\vspace{-1mm}

Given the Lab-Data in the base training, the trained inference model does not generalize well on test data collected under uncontrolled conditions, if not provided in the base training. The stacked bidirectional LSTM network failed to cope with data from real-world environments, as both data distributions differ too much. According to the baseline result provided in Table \ref{tab:retrain}, in most of the cases the accuracy equals to a uniform distribution. The model performs no better than random guessing on the Uncontrolled-Data. In our application, the total number of classes is nine, random guessing corresponds to an average accuracy of $\frac{1}{9}=11.1\,\%$ .

\vspace{-1.5mm}
\subsection{Our proposed approaches}
\vspace{-1mm}

\paragraph{The results of using the domain adaptation (DA) method} 
are depicted in Table \ref{tab:retrain}. Here we successively increased the amount of target labels from the subject development set to be included into the training to improve the performance of the domain adaptation. Without including any target labels, the network only minimizes the distribution of the source and target embeddings in an unsupervised way. The performance is only 10-20 percentage points better compared to the baseline. By incorporating more label information from the target domain, the knowledge transfer improves on the target domain classification. With only 50\,\% of the target labels, the results increased about 40 percentage points and with 100\,\% of the labels, the results increased about two to six folds in average.

\paragraph{Results of Few-shot Classification Networks}
Here, we evaluated three alternatives of the few-shot classification approach. These models typically do not need to retrain the pre-trained inference model, as the tasks are disjoint. An overview of the evaluation results is given in Table \ref{tab:no_retrain}. A general tendency is that the classification accuracy increases at least 5 percentage points with increasing number of support samples per class used in the evaluation. This is intuitive, due to an increased reliability and an improved decision boundary related to more support samples. The Siamese network (F1) provides similar results compared to the prototypical network (F2), as both network architectures work with euclidean similarity scores on the sample-base. Their performances are increased around 50 percentage points compared to the baseline model. The image-to-class measure in the LocalNet (F3) performs the best as depicted in Table \ref{tab:no_retrain} with an increase of two to six folds in average compared to the baseline. 

\begin{table*}[]
    \centering
    \begin{tabular}{l|l|c|c|c|c|c|c|c|c|c|c}
    \hline
         No. & Method & P1 & P2 & P3 & P4 & P5 &
            P1 & P2 & P3 & P4 & P5\\
         \hline
         - & Baseline & 16.25\,\% & 46.35\,\% & 11.17\,\% & 25.73\,\% & 15.5\,\% &
         -&-&-&-&-\\
         \hline
          &  & \multicolumn{5}{c|}{5 support samples each class} & 
          \multicolumn{5}{c}{10 support samples each class} \\
         \hline
          F1 & Siamese & 65.53\,\% & 83.77\,\% & 72.81\,\% & 65.61\,\% & 68.33\,\% &
          69.69\,\% & 90.07\,\% & 75.65\,\% & 69.08\,\% & 72.56\,\% \\
        \hline
         F2 & ProtoNet & 67.06\,\% & 79.9\,\% & 67.85\,\% & 77.54\,\% & 65.92\,\% &
         74.18\,\% & 88.59\,\% & 69.31\,\% & 87.88\,\% & 70.80\,\% \\
          \hline
         F3 & LocalNet & 85.28\,\% & 79.9\,\% & 64.85\,\% & 94.33\,\% & 86.98\,\% &
         \textbf{89.55\,\%} & \textbf{97.84\,\%} & \textbf{67.51\,\%} & \textbf{98.0\,\%} & \textbf{91.63\,\%} \\
    \hline
    \end{tabular}
    \caption{The accuracy for the three alternatives of few-shot classification task is shown. The number (5 or 10) indicates the number of support samples each class used in the evaluation. The term $P_i$ indicates the ID of the participants. The performance increases in general with the increasing number of supports. The LocalNet with the cosine similarity measure outperforms the other methods, as it includes the image-to-class feature correlation.}
    \label{tab:no_retrain}
    \vspace{-3mm}
\end{table*}

\subsection{Discussion}
\label{ch:discussion}
The performance of activity recognition based on ultrasound sensing using a mobile device is subject to many variables. The same hardware applied under changed conditions for the same person show variability in the signal strength. To deploy a fixed application to real-world scenarios is therefore not easy and often has to overcome some difficulties. In many cases, the performance drops due to the dissimilarity in both domains. To overcome this issue, we investigated several methods in this paper. We provide a database collected under various conditions to allow researchers perform experiments on it to solve the problem of lack of generalization on new individuals. The base data consists of Lab-Data under controlled environment and same sensing device. Uncontrolled setups from five different individuals are used to evaluate the methods for finetuning on individual dataset. 

Finetuning a base model on new domains requires sufficient amount of labeled samples from the Uncontrolled-Data. Otherwise, the model would overfit adopting on this small data amount. However, labels are most difficult to acquire and the individual labeling process might be error prone. In such cases, domain adaption method can be leveraged, where no label information of the target domain is required. Though, such network could benefit from including a small amount of the target labels in case both domains differ as investigated in our use-cases. 

Few-shot classification is suitable for adopting finetuning on limited data without retraining. This method can cope with individual hardware characteristics without modifying the base training. By comparing knowledge extracted from support samples of different categories, an unknown sample is able to assign to the correct category under the assumption that samples of similar categories are close in the embedding space. As no feature adaptation from the target domain is applied, this model requires both domains behave similarly. 

To leverage few-shot classification, the user has to pre-label a small amount of individual sessions before the model is adopted to this user. These labels are used to classify the new samples based on certain distance metrics. The developer does not need to modify the feature extraction network to individually adapt to each new user. In case of the domain adaptation, the developer needs to modify the base feature extraction according to the user data. It further assumes the similarity of both domains in order to avoid negative knowledge transfer.    

\section{Conclusion}
\label{ch:conclusion}
In this paper, we investigated different approaches to improve the generalizability of pre-trained classification models under controlled condition in uncontrolled real-world scenarios based on a mobile application for workout exercise recognition. We first presented a database to enable us analysing this problem and building novel solutions. This database allows us to study the body motion in relation to Doppler profiles from built-in hardware of different commercial smartphones. The gap between the development setup and the real-world scenarios, as we prove, often lead to performance drop and bad usability. The reason can be manifold, as it could rely on individual difference, hardware specifics or environmental changes. Our database is built to overcome this gap. 

We proposed two methods: domain adaptation and few-shot classification, to resolve the issue of lack of model generalizability. Our evaluations showed that the baseline method fails when faced with realistic data. Our proposed concept of using domain adaption without including the target labels improved the baseline only by 10-20 percentage points in most cases. However, this method benefits from including target labels, as with increasing amount of target labels in the training phase, the recognition performance increases by two to six folds compared to the baseline. Our proposed solution that is based on few-shot classification improved the accuracy to the same range, however, without the effort of retraining.

\vspace{-1.5mm}
\section*{Acknowledgment}
This research work has been partially funded by the German Federal Ministry of Education and Research and the Hessen State Ministry for Higher Education, Research and the Arts within their joint support of the National Research Center for Applied Cybersecurity ATHENE
\vspace{-1.5mm}



%

\bibliographystyle{IEEEtran}
\bibliography{IEEEexample}
\end{document}